\documentclass[letterpaper, 10 pt, conference]{ieeeconf}  

\IEEEoverridecommandlockouts

\usepackage[T1]{fontenc}
\usepackage{caption}
\usepackage{cite}

\usepackage{amssymb}
\usepackage{graphicx}
\usepackage{booktabs}
\usepackage{algorithm}
\usepackage{algorithmic}
\usepackage{multirow}
\usepackage{amsmath}
\usepackage{wasysym}
\usepackage{bbm}
\usepackage{mathtools}
\usepackage{tablefootnote}
\usepackage{pifont}
\usepackage[utf8]{inputenc}
\usepackage{float}
\usepackage{placeins}

\title{3D Dynamics-Aware Manipulation: Endowing Manipulation Policies with 3D Foresight}

\author{Yuxin He$^{1}$, Ruihao Zhang$^{1 \dagger}$, Xianzu Wu$^{1 \dagger}$, Zhiyuan Zhang$^{1}$, Cheng Ding$^{2}$, Qiang Nie$^{1 \star}$
\thanks{$^{1}$The Hong Kong University of Science and Technology (Guangzhou).}
\thanks{$^{\dagger}$Work done during internship. \quad $^{2}$JAKA Robotics Co., Ltd.}
}

\begin{document}

\maketitle
\thispagestyle{empty}
\pagestyle{empty}

\begin{abstract}

    The incorporation of world modeling into manipulation policy learning has pushed the boundary of manipulation performance. However, existing efforts simply model the 2D visual dynamics, which is insufficient for robust manipulation when target tasks involve prominent depth-wise movement. To address this, we present a 3D dynamics-aware manipulation framework that seamlessly integrates 3D world modeling and policy learning. Three self-supervised learning tasks (current depth estimation, future RGB-D prediction, 3D flow prediction) are introduced within our framework, which complement each other and endow the policy model with 3D foresight. Extensive experiments on simulation and the real world show that 3D foresight can greatly boost the performance of manipulation policies without sacrificing inference speed. Code is available at https://github.com/Stardust-hyx/3D-Foresight.

\end{abstract}

\section{INTRODUCTION}

An exciting direction for improving language-conditioned manipulation policies is the incorporation of world modeling, i.e., predicting the transition of world states driven by impetus like language command or low-level actions. Recently, this line of work has demonstrated promising results by pretraining policy models on large-scale video data to predict future RGB observation \cite{wuunleashing,li2024gr,tian2024predictive,Zhang2025UPVLAAU,wang2025unified}. Through this kind of 2D world model learning, policy models become aware of the desired future states under current observations and can more easily predict actions that lead to such states.

Unfortunately, the monocular 2D description of the world is lossy in terms of depth information, which is valuable for distance guidance and obstacle avoidance. Yet \textbf{it is possible to infer depth well from a monocular image}, as showcased by one-eyed persons or monocular depth estimation neural networks \cite{yang2024depth,bochkovskii2024depth,chen2025video}. This leads to a key insight behind our work: Instead of praying for our models to develop such an ability implicitly, it is more practical to explicitly teach our models about that. Another key insight is that \textbf{both the 3D scene transformation and low-level SE(3) robotic actions share the same 3D space and a similar dynamics}, i.e., the underlying trend of how everything should move driven by the same language command. Policy models with 3D foresight will be able to capture this underlying trend and behave accordingly.

Motivated by these two insights, we come up with a 3D dynamics-aware manipulation framework that seamlessly integrates 3D world modeling and policy learning, so as to endow manipulation policies with 3D foresight. In the core of our framework are three complementary self-supervised learning tasks: current depth estimation, future RGB-D prediction, and 3D flow prediction. Cross-embodiment pretraining and downstream fine-tuning are conducted with these auxiliary learning objectives.

\begin{figure}
    \centering
    \includegraphics[width=1.0\columnwidth]{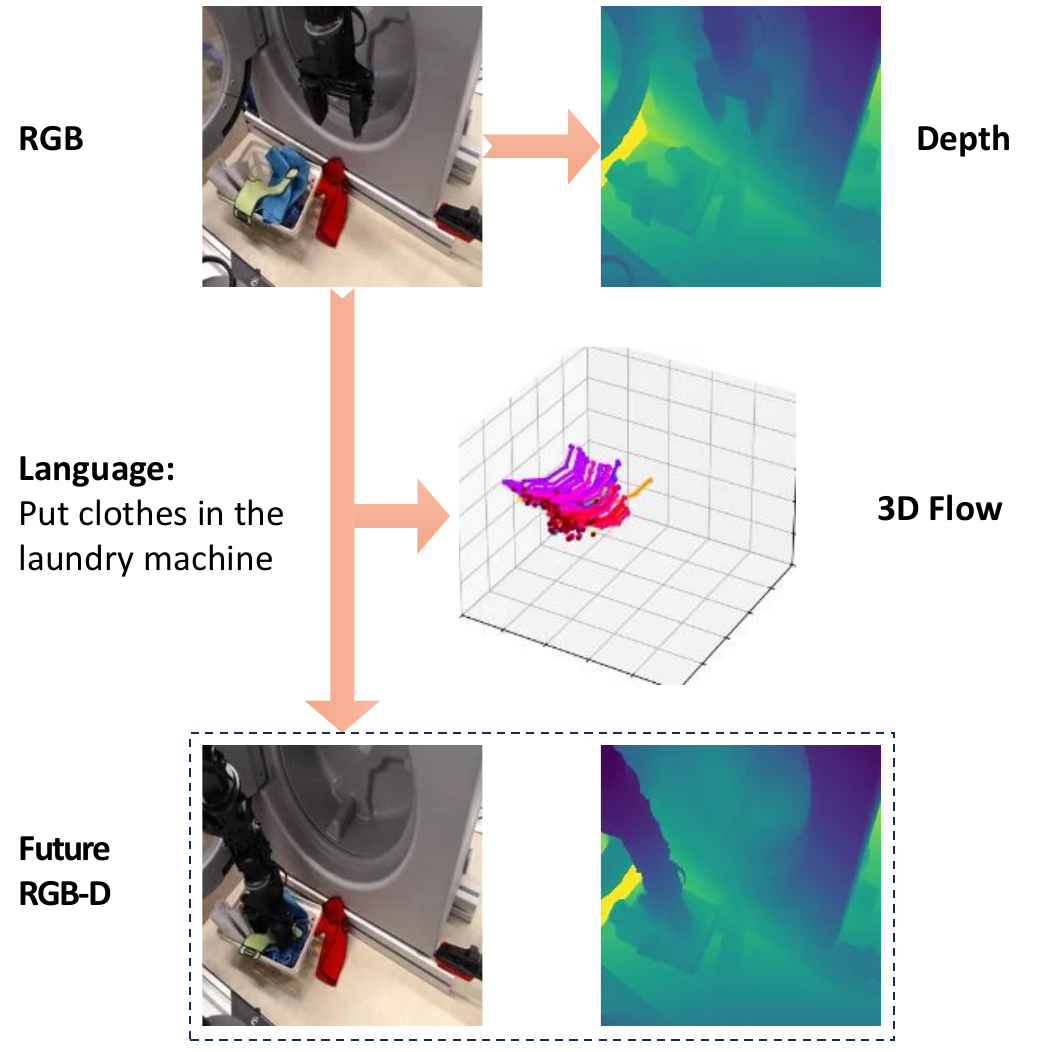}
    \caption{We propose to equip manipulation policies with 3D foresight by letting the policy models learn to predict depth, future RGB-D and 3D flow from observed RGB. We lift large-scale manipulation demonstration videos into 3D and track the points within to achieve self-supervised pretraining and finetuning.}
   \label{fig:intro}
   \vspace{-1em}
\end{figure}

We choose to represent the 3D world with RGB-D to reduce the burden of data preprocessing. Although it is possible to further lift videos into sequences of point clouds via 3D reconstruction, how to satisfactorily achieve that on large-scale in-the-wild data with limited computing resources is still an open question, which is why we leave it for future research. In order to represent the transformation of 3D scene, we leverage 3D flow \cite{wen2023any,yuan2024general, chen2024g3flow} as a bridge between current and future RGB-D frames, as shown in Fig. \ref{fig:intro}. We find that combining these related factors can let the model learn to calibrate each other during the learning process.

A causal transformer is employed to jointly model the language-driven dynamics of RGB-D, 3D flow and SE(3) robotic actions in an end-to-end manner, maximizing parameter sharing and knowledge transfer. To avoid unnecessary latency, the transformer applies a query-based parallel representation updating mechanism. For non-action output, we utilize auxiliary decoding heads to compute corresponding losses during training and remove or offload these auxiliary heads during inference.

Our experiments are carried out on two simulation benchmarks (CALVIN \cite{mees2022calvin}, LIBERO \cite{liu2024libero}) as well as real-world settings, where policy models equipped with 3D foresight achieve state-of-the-art (SoTA) without sacrificing inference speed. In-depth analyses reveal the contribution of 3D foresight to handling tasks that involve prominent depth-wise movement.

In summary, the main contributions of this paper include:
\begin{itemize}
    \item We propose to endow manipulation policies with 3D foresight by combining 3D world modeling and policy learning under a unified framework.
    \item Three self-supervised learning tasks (current depth estimation, future RGB-D prediction, 3D flow prediction) are introduced to capture 3D world dynamics.
    \item The benefit of 3D foresight is verified by experiments on two simulation benchmarks and the real world.
\end{itemize}

\section{RELATED WORK}

\subsection{Language-conditioned Manipulation}
Instead of constructing a single-task visuomotor policy for every task, it is more desirable to have a multitask language-conditioned manipulation policy that is generalizable. CLIPort \cite{shridhar2022cliport}, which injects frozen language embeddings into two-stream convolution networks, is one of the first studies that show the potential of this paradigm. However, purely relying on frozen language embeddings will impose a limit of generalizability, due to the domain gap between language and vision. Many methods \cite{10160591,brohan2023rt,kim2024openvla,black2410pi0,he2026towards} leverage large vision language models (VLMs) to address this. Despite the promising results that these methods have achieved, they face efficiency and interpretability issues. Another line of work \cite{wuunleashing,li2024gr,tian2024predictive,Zhang2025UPVLAAU,wang2025unified} tries to mine the dynamics hidden in captioned videos, which is known as the world modeling approach \cite{ha2018recurrent}.

\subsection{World Modeling for Manipulation Policy Learning}

Similar to VLM-based methods, the world modeling approach benefits from self-supervised learning on easily accessible unlabeled data. But the world modeling approach is more interpretable, as the core of it is to predict the outcome of behaviors. Existing work \cite{wuunleashing,li2024gr,tian2024predictive,Zhang2025UPVLAAU,wang2025unified} mainly leverages future image generation as the self-supervised learning objective. However, the dynamics of pixels in the 2D frame space is superficial and just loosely related to the dynamics of robotic actions. Our work highlights the significance of modeling the dynamics in 3D space.

\subsection{Flow-enhanced Manipulation Polices}
Different from modeling frame-to-frame transition, flow highlights a subset of points within the scene. As a result, movement information is decoupled from visual appearance information, which makes it easier to capture by a model. In \cite{bharadhwaj2024gen2act}, 2D flow prediction from the latent space of the policy model acts as an auxiliary learning task, while the predicted flow is not utilized for action guidance. In contrast, all other flow-enhanced manipulation policies \cite{vecerik2024robotap, wen2023any, bharadhwaj2024track2act, xuflow} solely employ predicted flow to guide action prediction. Among them, flow is typically confined to the 2D frame space, except for \emph{General Flow} \cite{yuan2024general} and \emph{G3Flow} \cite{chen2024g3flow}, which also adopt 3D flow. However, the focus of \emph{General Flow} is developing a 3D flow prediction model and the focus of \emph{G3Flow} is semantic-oriented 3D reconstruction and representation, whereas our work features end-to-end integration of 3D flow.

\begin{figure*}
    \centering
    \includegraphics[width=1.0\linewidth]{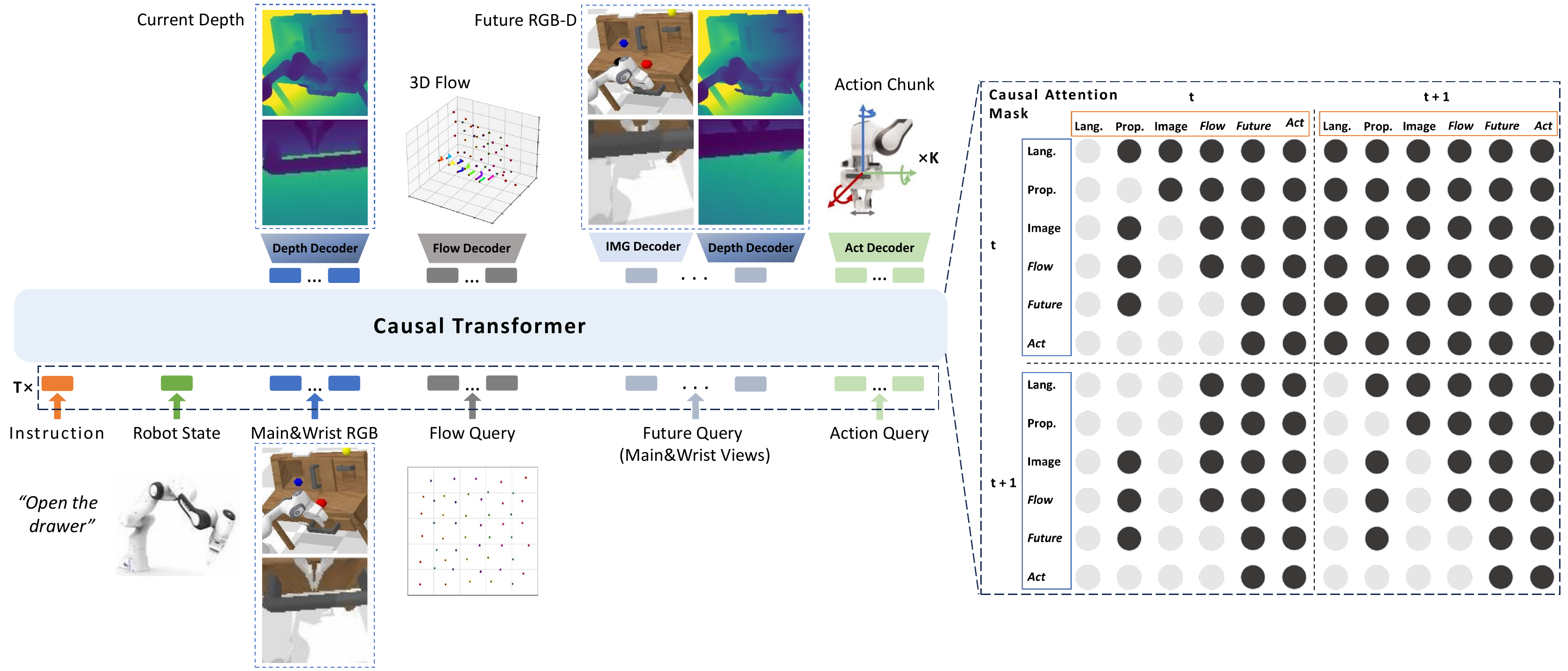}
    \caption{An overview of the proposed end-to-end framework that captures the language-driven dynamics of 3D world and low-level manipulation actions in a unified manner. In the illustration of the causal attention mask, a light circle mean the horizontal element can attend to the vertical element.}
    \label{fig:model}
\end{figure*}

\section{METHOD}

A language-conditioned action-chunking policy $\pi_\theta(\mathbf{a}_{t:t+K-1} | \mathbf{o}_{t-T+1:t}, \mathbf{c})$ takes a language command $\mathbf{c}$ and observations $\mathbf{o}_{t-T+1:t}$ (images, proprioception states, etc.) with historical window size $T$ as input, and outputs a chunk of actions $\mathbf{a}_{t:t+K-1}$ of length $K$. These variables belong to quite distinct modalities, which is why \textbf{knowledge from large-scale diverse data and auxiliary learning objectives that narrow the domain gap are necessary}. 3D Foresight incorporates the prediction of current depth $\mathbf{d}^\text{main/wrist}_{t}$, future RGB-D $\mathbf{rgbd}^\text{main/wrist}_{t+S}$ in main/wrist views and 3D flow $\boldsymbol{\tau}_{t:t+L-1} \in \mathbb{R}^{L \times P \times 3}$, where $S$ is the time shift of future images, $L$ is the flow length and $P$ is the number of track points. In our formulation, the three dimensions of 3D flow correspond to x, y (in pixel coordinates) and metric depth value, respectively. Our framework employs a causal transformer to model the multimodal spatial-temporal correlation, as shown in Fig. \ref{fig:model}.

\subsection{Causal Modeling of Multi-modal Input and Queries}

Following previous research \cite{wuunleashing, li2024gr}, the language command is encoded into a vector $\mathbf{c} \in \mathbb{R}^d$ ($d$ is the model dimension) with CLIP \cite{radford2021learning} text encoder and a linear projection layer. Each main/wrist-view RGB image is encoded into a matrix of $(1+r)$ vectors via MAE \cite{he2022masked} and a perceiver resampler, where the first vector is the \emph{CLS} token and the rest $r$ vectors are resampled from patch tokens. The robot proprioception state, which includes the 6D end effector pose and the binary gripper status, is embedded into a vector $\mathbf{p} \in \mathbb{R}^d$ with linear layers.

The construction of queries for 3D flow, future RGB-D and action chunk is introduced below: 

\paragraph{Flow Query} The model predicts the future 3D trajectories of grid points (during inference) or randomly sampled points near grids (during training). We initialize $l$ learnable vectors as the flow query. To include information about which points to track, the starting pixel coordinates of sampled points are encoded into a vector via a linear layer, which is added to each flow query vector.

\paragraph{Future Query} The query for future RGB-D in main/wrist-view is instantiated as $1+r$ learnable vectors. The number of future query vectors is equal to the number of image embedding vectors such that appropriate capacity is equipped to reconstruct future RGB-D.

\paragraph{Action Query} A learnable vector is responsible for querying action-related information.

The language, vision, proprioception input along with various queries for each timestep are organized into a sequence. And all sequences from timestep $t-T+1$ to timestep $t$ are concatenated and fed into a GPT-style transformer equipped with a carefully designed self-attention mask \textbf{in one pass}. Within it, all language, vision, proprioception tokens attend to their historical counterparts to establish temporal relation. All queries attend to current and historical language, vision tokens. Action query additionally attends to current and historical proprioception tokens as well as 3D flow query, since the two are deeply correlated. The hidden states of all tokens are updated by causal self-attention in parallel. Please refer to Fig. \ref{fig:model} for an illustration of the attention mask.

\subsection{3D World Model Learning and Policy Learning}

Three complementary self-supervised learning objectives equip the model with 3D foresight:

\subsubsection{Current Depth Prediction}
We instantiate a depth decoder based on bidirectional self-attention. The input to the depth decoder includes the linear projection of the final hidden states of current main/wrist-view image tokens (acting as context) and a set of 2D masked patch tokens. Here, the masked patch tokens are the sum of 2D sin-cos positional embeddings and a learnable mask vector. The final hidden states of these masked patch tokens are linearly transformed into the predicted current depth.

\subsubsection{Future RGB-D Prediction}
The final representation of future query goes into an image decoder as well as the depth decoder. The image decoder is structurally identical to the depth decoder, except that its output has three channels rather than one channel.

\subsubsection{3D Flow Prediction.}
The flow decoder works in a similar way as the depth decoder. It first initiates a set of masked patch tokens from starting pixel coordinates, then updates the patch representation conditioned on the final representation of flow query via bidirectional self-attention, then linearly transforms the final patch representation into predicted flow $\hat{\boldsymbol{\tau}}_{t:t+L-1} \in \mathbb{R}^{L \times P \times 3}$.

The model also learns to predict the action chunk in an imitation manner. As in ACT \cite{zhao2023learning} and GR-MG \cite{wuunleashing}, a CVAE encoder compresses the ground-truth action chunk into a latent vector, which are fed into a transformer decoder, along with the final representation of action query and a sequence of positionally-embedded mask tokens. The final hidden states of these mask tokens are transformed into the predicted action chunk $\hat{\mathbf{a}}_{t:t+K-1}$ through an MLP layer. In this paper, we adopt the widely-used SE(3) action space, where an action includes translation (delta of $x, y, z$), rotation (delta of roll, pitch, yaw) and target binary closeness of the end effector.

The end-to-end learning loss $\mathcal{L}$ over a timestep frame is the combination of prediction losses over current depth, future RGB-D, 3D flow and action chunk:
\begin{align}
&\mathcal{L}_\text{depth} = \sum_{v \in \{\text{main}, \text{wrist}\}} \mathrm{MSE}(\hat{\mathbf{d}}^{v}_t, \tilde{\mathbf{d}}^{v}_t) \\
&\mathcal{L}_\text{future} = \sum_{v \in \{\text{main}, \text{wrist}\}} \mathrm{MSE}(\hat{\mathbf{rgbd}}^{v}_{t+S}, \tilde{\mathbf{rgbd}}^{v}_{t+S}) \\
&\mathcal{L}_\text{flow} = \mathrm{MSE}(\hat{\boldsymbol{\tau}}_{t:t+L-1}, \boldsymbol{\tau}_{t:t+L-1}) \\
&\mathcal{L}_\text{act} = \mathrm{SmoothL1}(\hat{\mathbf{a}}_{t:t+K-1}[:6], {\mathbf{a}}_{t:t+K-1}[:6]) \nonumber\\
&\quad \quad + \alpha \cdot \mathrm{BCE}(\hat{\mathbf{a}}_{t:t+K-1}[6], {\mathbf{a}}_{t:t+K-1}[6]) \\
&\mathcal{L} = \beta \cdot \mathcal{L}_\text{depth} + \gamma \cdot \mathcal{L}_\text{future} + \delta \cdot \mathcal{L}_\text{flow} + \mathcal{L}_\text{act}
\end{align}
where $\mathrm{MSE}$, $\mathrm{SmoothL1}$ and $\mathrm{BCE}$ mean Mean Squared Error, Smooth L1 and Binary Cross Entropy respectively. $\tilde{\mathbf{rgbd}}^{v}$ and $\tilde{\mathbf{d}}^{v}$ mean that the pixel-wise normalization \cite{he2022masked} is applied to the target value of the R, G, B and depth channels. We set $\alpha=0.01$ following GR-MG\cite{li2024gr}, and $\beta=0.05$, $\gamma=0.1$, $\delta=0.1$, which work well consistently throughout three environments (CALVIN, LIBERO, and the real world).

\begin{figure*}[t]
    \centering
    \includegraphics[width=0.95\linewidth]{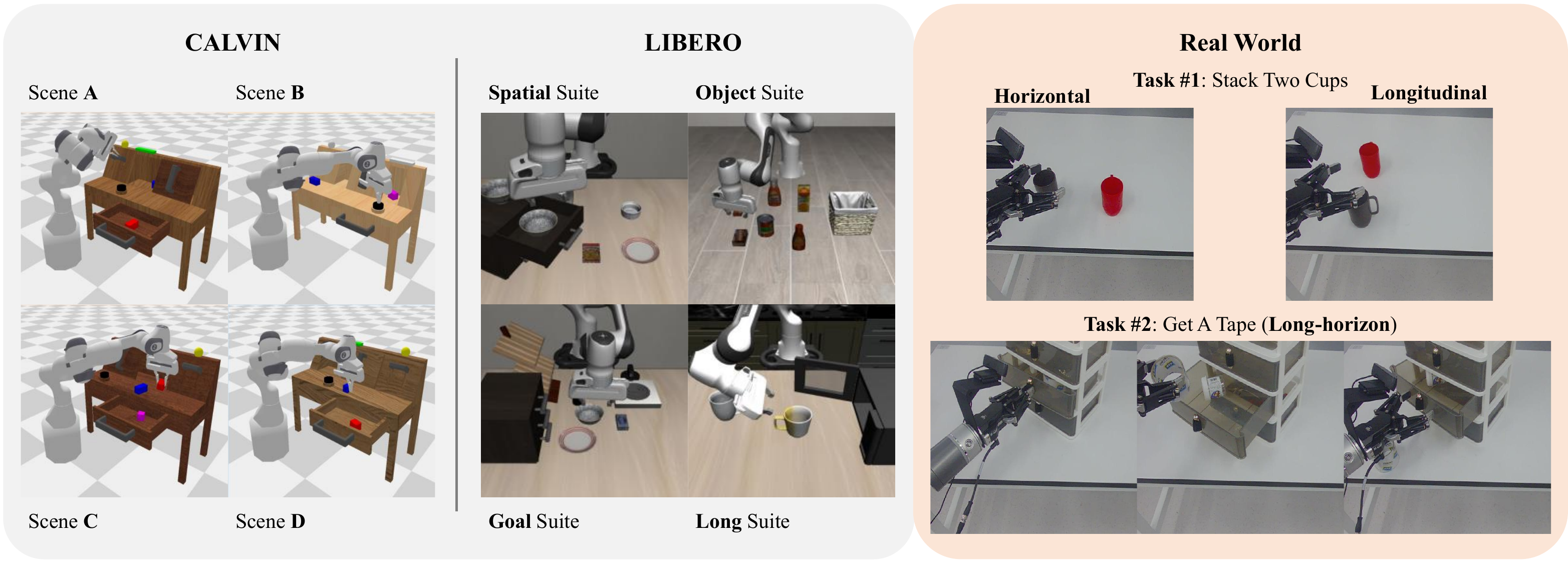}
    \caption{Environments for our experiments. CALVIN involves 4 scenes of different colors, textures and object placements and 34 manipulation tasks. LIBERO features 4 evaluation task suites that challenge different dimensions of capability. Our real-world setups involve two tasks that require strong spatial awareness.}
    \label{fig:env}
\end{figure*}

\subsection{Automatic Annotation of Depth and 3D Flow}
\label{ann_depth_flow}

In many datasets, depth information is not recorded and we use a combination of SoTA monocular metric depth estimator Depth-Anything-V2 \cite{yang2024depth} and video depth estimator Video-Depth-Anything \cite{chen2025video} to obtain temporally consistent metric depth estimate. For any RGB-D video, DELTA \cite{ngo2024delta} can efficiently track the 3D position of points within. We employ a sampling strategy that results in nearly 1250 track points with $\frac{1}{4}$ of them moving in the video and the others staying still. We do so to maintain a trade-off balance between teaching the model the 3D dynamics and narrowing the training-inference gap (since track points are just uniformly sampled from grids during inference).

\subsection{Cross-embodiment Pretraining}
We pretrain our model on large-scale cross-embodiment manipulation video data. During pretraining, proprioception states are excluded from the input; actions are excluded from the output; wrist-view depth/images are excluded from the input and output, since these elements are always heterogeneous or missing in different data sources.

\subsection{Implementation Details}
We instantiate our models with GR-MG \cite{li2024gr} checkpoint pretrained on Ego4D \cite{grauman2022ego4d} videos. GR-MG is a representative policy model based on 2D world modeling, which shares a similar architecture as ours, except for the depth and 3D flow related parts. Note that, GR-MG additionally consists of a goal image generation module based on image editing, and we keep it for a fair comparison with GR-MG. We set the historical window size $T$ as 10, the future interval $S$ as 3, the action chunk size $K$ as 5 and the length of 3D flow $L$ as 6. Our pretraining is then conducted on 44K trajectories from 5 datasets (RH20T \cite{fang2023rh20t}, Bridge \cite{walke2023bridgedata}, Berkeley UR5 \cite{BerkeleyUR5Website} , Mutex \cite{shah2023mutex} and LIBERO \cite{liu2024libero}) that cover 4 kinds of embodiments (humans, Franka, UR5, WidowX robots). The pretraining process lasts for 35 epochs, which takes 3 days on 4 NVIDIA 4090 GPUs. After that, we finetune and evaluate our models on downstream manipulation tasks. 

\section{EXPERIMENTS}

Our experiments aim to answer the following questions:
\begin{itemize}
    \item \textbf{Q1:} Can policy models benefit from 3D foresight?
    \item \textbf{Q2:} Is 3D foresight superior to 2D foresight?
    \item \textbf{Q3:} How do different learning objectives contribute to performance? Are these objectives complementary?
    \item \textbf{Q4:} Does the proposed method work in the real world?
\end{itemize}

\subsection{Setup}

\subsubsection{Environments}
Our simulation experiments are carried out on two benchmarks (CALVIN, LIBERO) and the real world, as shown in Fig. \ref{fig:env}. CALVIN \cite{mees2022calvin} encompasses 34 manipulation tasks and 4 different scenes (A, B, C, D). 5K expert trajectories with language instructions are provided for each scene. LIBERO \cite{liu2024libero} consists of 5 task suites (Spatial, Object, Goal, Long and 90). Each task suite has 10 tasks, except for the last task suite, which has 90 tasks. 50 demonstrations are provided for each task. Our real-world experiments are carried out on a JAKA K-1 7DoF robotic arm with a gripper, a fixed main camera (Orbber Gemini 2L) and a wrist camera (Logitech C922Pro). Two tasks that involve prominent depth-wise movement are considered, including: 1) stack two cups; 2) open a drawer, pick a tape from the drawer and place it on the table then close the drawer. We collect 60 demonstrations for each task using a VR-based teleoperation system.

Ground-truth depth values are available in CALVIN, which are directly used as depth labels during finetuning; in LIBERO and our real-world settings, ground-truth depth values are inaccessible and we obtain the depth labels for training with our preprocessing pipeline. 

\begin{table*}[t]
    \centering
    \caption{Overall performance comparison on CALVIN. During evaluation, 1000 chains of tasks are randomly sampled, and the success rates of consecutive 5 tasks are recorded and averaged over all chains. ``1'' $\sim$ ``5'' indicate the average success rates of completing 1 $\sim$ 5 tasks. ``Avg. Len.'' means the average number of completed tasks.}
    \label{calvin_result}
    \scalebox{0.96}[0.96]{
	\begin{tabular}{l|cccccccccccc}
		\toprule
        \multirow{2}{*}{\textbf{Method}} & \multicolumn{6}{c}{\textbf{D $\rightarrow$ D}} & \multicolumn{6}{c}{\textbf{ABC $\rightarrow$ D}} \\
        \cmidrule(lr){2-7} \cmidrule(lr){8-13} & {1} & {2} & {3} & {4} & {5} & 
        \textbf{Avg. Len.} & {1} & 2 & {3} & {4} & {5} & 
        \textbf{Avg. Len.} \\
        \midrule
        3D Diffusion Actor \cite{ke20243d} & - & - & - & - & - & - & 93.8 & 80.3 & 66.2 & 53.3 & 41.2 & 3.35 \\
        RoboUniView \cite{liu2024robouniview} & 96.2 & 88.8 & 77.6 & 66.6 &  56.3 & 3.85 & 94.2 & 84.2 & 73.4 & 62.2 & 50.7 & 3.64 \\
        GR-1 \cite{wuunleashing} & - & - & - & - & - & - & 85.4 & 71.2 & 59.6 & 49.7 & 40.1 & 3.06 \\
        SeeR (base) \cite{tian2024predictive} & - & - & - & - & - & - & 94.4 & 87.2 & 79.9 & 72.2 & 64.3 & 3.98 \\
        UP-VLA \cite{Zhang2025UPVLAAU} & - & - & - & - & - & - & 92.8 & 86.5 & 81.5 & 76.9 & 69.9 & 4.08 \\
        GR-MG \cite{li2024gr} & 93.0 & 84.5 & 76.5 & 69.0 & 60.8 & 3.84 & 96.8 & 89.3 & 81.5 & 72.7 & 64.4 & 4.04 \\
        \midrule
        2D Foresight (scratch) & 94.8 & 85.1 & 76.6 & 70.8 & 62.5 & 3.90 & 95.6 & 90.5 & 83.1 & 74.8 & 64.2 & 4.08 \\
        3D Foresight (scratch) & 95.5 & 87.6 & 80.9 & 73.2 & 64.1 & 4.01 & 96.2 & 91.1 & 84.4 & 77.1 & 68.7 & 4.15 \\
        3D Foresight & 95.7 & 88.0 & 81.6 & 74.9 & 66.3 & 4.08 & 96.9 & 92.0 & 85.7 & 78.8 & 71.3 & 4.23 \\
		\bottomrule
	\end{tabular}
    }
\end{table*}

\begin{table}[ht]
    \centering
    \caption{Overall Performance Comparison on LIBERO}
    \label{libero_result}
    \scalebox{0.92}[0.92]{
	\begin{tabular}{l|ccccc}
		\toprule
        \textbf{Method} & \textbf{Spatial} & \textbf{Object} & \textbf{Goal} & \textbf{Long} & \textbf{Avg. SR} \\
        \midrule
        ATM \cite{wen2023any} & 84.0 & 89.4 & 79.6 & 65.2 & 79.6 \\
        GR-1 \cite{wuunleashing} & 93.4 & 93.4 & 89.0 & 84.2 & 90.1 \\
        GR-MG \cite{li2024gr} & 94.0 & 94.8 & 91.6 & 86.4 & 91.7 \\
        \midrule
        2D Foresight (scratch) & 94.4 & 95.8 & 93.0 & 87.2 & 92.6 \\
        3D Foresight (scratch) & 95.8 & 97.4 & 94.0 & 90.2 & 94.3 \\
        3D Foresight & 96.4 & 98.0 & 94.8 & 92.0 & 95.3 \\
		\bottomrule
	\end{tabular}
    }
\end{table}

\subsubsection{Baselines}
We compare our method with the following baselines that \emph{leverage world modeling or 3D information}:
\begin{itemize}
    \item \textbf{3D Diffusion Actor} \cite{ke20243d} enhances Diffusion Policy with 3D scene representations.
    \item \textbf{RoboUniView} \cite{liu2024robouniview} enhances RoboFlamingo with view-invariant 3D representations based on 3D occupancy.
    \item \textbf{ATM} \cite{wen2023any} incorporates 2D flow predicted by a track transformer to guide a transformer policy.
    \item \textbf{GR-1} \cite{wuunleashing} is a GPT-style policy pretrained with the future RGB prediction task on large-scale video data.
    \item \textbf{SeeR} \cite{tian2024predictive} enhances GR-1 by increasing the model size and combining video pretraining with inverse dynamics.
    \item \textbf{GR-MG} \cite{li2024gr} enhances GR-1 with action chunking and goal images generated by an image editing model.
    \item \textbf{UP-VLA} \cite{Zhang2025UPVLAAU} a VLA model that combines multimodal understanding and future RGB prediction.
    \item \textbf{2D Foresight (scratch)} is a 2D counterpart of our method that enhances GR-MG by combining future RGB prediction with 2D flow (without cross-embodiment pretraining).
    \item \textbf{3D Foresight (scratch)} enhances GR-MG with the proposed 3D world modeling objectives (without cross-embodiment pretraining).
\end{itemize}

We do not compare with UniVLA\cite{wang2025unified} and DreamVLA\cite{dreamvla25} because they are based on much (at least 5$\times$) larger backbones and it is unfair to compare with them.

\begin{figure}
    \centering
    \includegraphics[width=1.0\columnwidth]{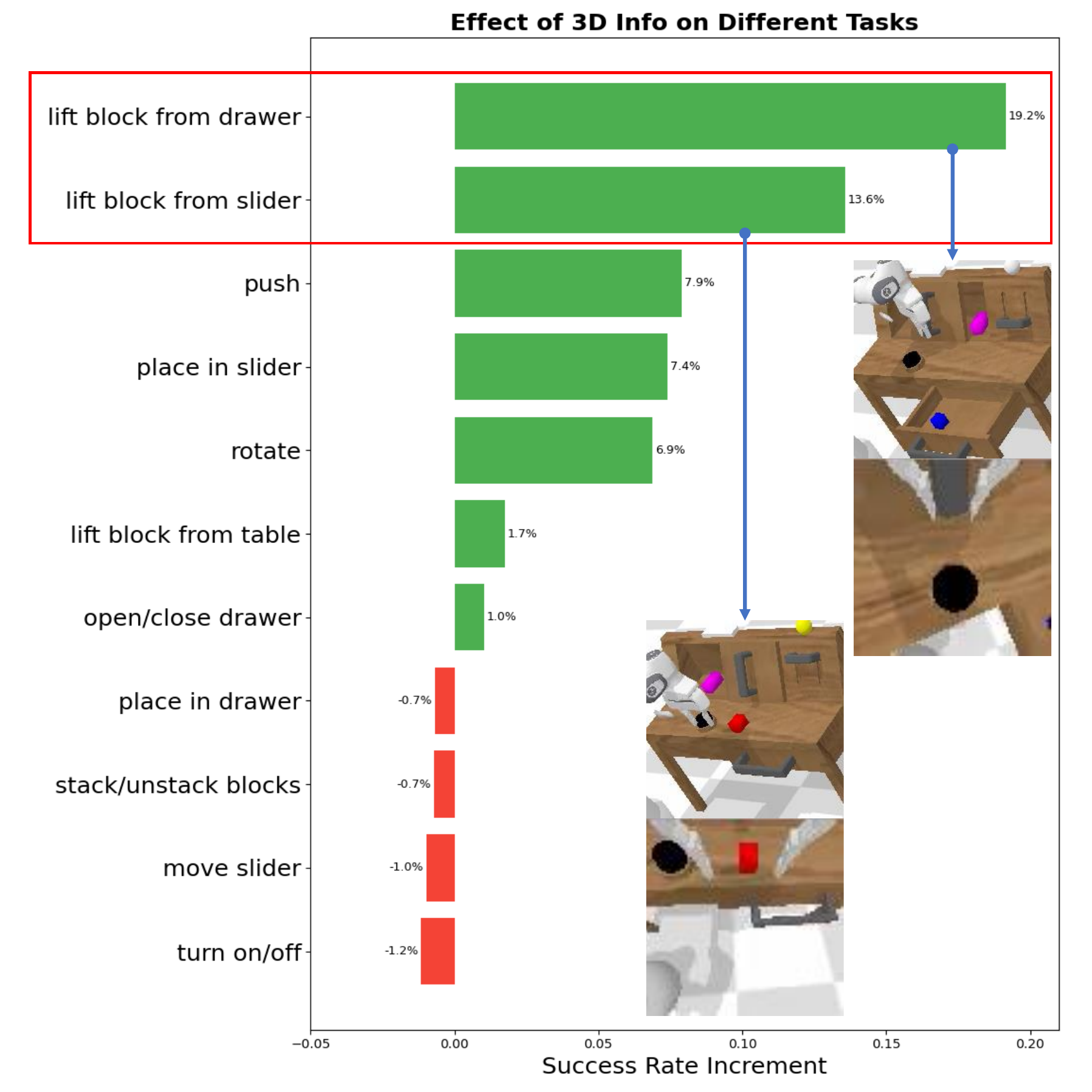}
    \caption{The increment of task-wise succuss rates on CALVIN D$\rightarrow$D after evolving from 2D Foresight to 3D Foresight.}
   \label{fig:analyze_increment}
\end{figure}

\begin{table*}[t]
\begin{minipage}{0.66\linewidth}
    \centering
    \captionof{table}{Ablation Study on CALVIN D$\rightarrow$D and the Real World}
    \label{ablation}
    \scalebox{0.91}[0.91]{
	\begin{tabular}{l|ccccccc}
		\toprule
        \multirow{2}{*}{\textbf{Method}} & \multicolumn{4}{c}{\textbf{CALVIN D $\rightarrow$ D}} & \multicolumn{3}{c}{\textbf{Real-World}} \\
        \cmidrule(lr){2-5} \cmidrule(lr){6-8} & $\mathcal{L}_\text{depth}$ & $\mathcal{L}_\text{future}$ & $\mathcal{L}_\text{flow}$ & \textbf{Avg. Len.} & $\mathcal{L}_\text{depth}$ & $\mathcal{L}_\text{future}$ & $\mathcal{L}_\text{flow}$ \\
        \midrule
        3D Foresight (scratch) & 0.021 & 0.11 & 6.8e-5 & 4.01 & 0.043 & 0.20 & 3.9e-4 \\
        \midrule
        w/o Current Depth & - & 0.12 & 7.5e-5 & 3.97 & - & 0.23 & 4.4e-4 \\
        w/o Future RGB-D & 0.027 & - & 8.4e-5 & 3.94 & 0.048 & - & 4.8e-4 \\
        w/o 3D Flow & 0.024 & 0.14 & - & 3.95 & 0.047 & 0.25 & - \\
		\bottomrule
	\end{tabular}
    }
\end{minipage}
\hfill 
\begin{minipage}{0.33\linewidth}
    \centering
    \captionof{table}{Inference Speed Comparison}
    \label{infer_speed}
    \scalebox{0.92}[0.92]{
	\begin{tabular}{lc}
		\toprule
        \textbf{Method} & \textbf{Inference Latency} \\
        \midrule
        RoboUniView \cite{liu2024robouniview} & 105 ms \\
        ATM \cite{wen2023any}  & 38 ms \\
        GR-1 \cite{wuunleashing} & 35 ms \\
        UP-VLA \cite{Zhang2025UPVLAAU} & 252 ms \\
        GR-MG \cite{li2024gr} & 106 ms \\
        3D Foresight & 112 ms (+6 ms) \\
		\bottomrule
	\end{tabular}
    }
\end{minipage}
\end{table*}

\begin{table}[t]
    \centering
    \caption{Performance Comparison on the Real World}
    \label{realwolrd_result}
    \scalebox{0.92}[0.92]{
	\begin{tabular}{l|ccc}
        \toprule
		\multirow{2}{*}{\textbf{Method}} & \multicolumn{2}{c}{Stack Two Cups} & \multirow{2}{*}{Get A Tape} \\
        \cmidrule(lr){2-3} & \textbf{Horizontal} & \textbf{Longitudinal} & \\
        \midrule
        ATM & 75\% & 50\% & 40\% \\
        GR-MG & 70\% & 40\% & 55\% \\
        \midrule
        2D Foresight (scratch) & 75\% & 35\% & 60\% \\
        3D Foresight (scratch) & 75\% & 60\% & 65\% \\
        3D Foresight & 80\% & 70\% & 75\% \\
		\bottomrule
	\end{tabular}
    }
\end{table}

\subsection{Can Policy Models Benefit from 3D Foresight? (Q1)}

Main results on CALVIN are shown in Table \ref{calvin_result}. Without pretraining, 3D foresight boosts the performance of GR-MG from 3.84 to 4.01 in the in-domain setting (D$\rightarrow$D) and from 4.04 to 4.15 in the zero-shot scene transfer setting (ABC$\rightarrow$D). This suggests that 3D foresight is directly beneficial for manipulation and the benefit will be greater if the target scene is seen during training. With pretraining, the model performance further increases by 0.07 in the in-domain setting and by 0.08 in the zero-shot scene transfer setting, demonstrating the advantage of our framework in leveraging cross-embodiment manipulation videos.

Table \ref{libero_result} displays the evaluation results on LIBERO. In contrast to CALVIN, LIBERO does not provide ground-truth depth values, and the depth labels for training are obtained with our preprocessing pipeline. Despite that, 3D foresight consistently increases the performance of GR-MG by large margins across the 4 diverse task suites with or without pretraining. This indicates the robustness of our method to noisy pseudo depth labels and the versatility of our method to different scenes that challenge different aspects of manipulation ability.

It is worth mentioning that 3D foresight provides all the benefits with only a negligible increase in inference cost. We compare the inference latency of the baseline methods and our method in Table \ref{infer_speed}. It turns out that our model is only 6 ms slower than the backbone GR-MG. This is achieved by removing or offloading the auxiliary decoding heads for current depth, future RGB-D and 3D flow during inference, since these prediction output are only the by-product of our self-supervise learning objects.

\subsection{2D Foresight vs. 3D Foresight (Q2)}

To compare the effect of 2D foresight and the effect of 3D foresight in a more rigorous way, we implement a strict 2D counterpart of our model. Concretely, we equip GR-MG with a 2D flow prediction module, where the depth dimension is ablated. According to Table \ref{calvin_result} and Table \ref{libero_result}, with the 2D flow prediction module, the performance of GR-MG  increases from 3.84 to 3.90 (+0.06) in CALVIN D$\rightarrow$D, from 4.04 to 4.08 in CALVIN ABC$\rightarrow$D (+0.04), and from 91.7 to 92.6 (+0.09) in LIBERO, but still lags far behind the proposed 3D foresight method by 0.11, 0.07 and 0.17. The results demonstrate that the advantage of our method does not simply root in the integration of flow prediction, but the comprehensive integration of 3D world modeling.

A more in-depth analysis of the superiority of 3D foresight over 2D foresight is carried out on CALVIN D$\rightarrow$D, where we calculate the success rate increments of different manipulation tasks. As shown in Fig. \ref{fig:analyze_increment}, two outstanding tasks that benefit the most from 3D foresight are ``lift block from drawer'' and ``lift block from slider'', both of which involve prominent depth-wise movement.

\subsection{Ablation Study (Q3)}

To investigate the contributions of different self-supervised learning objectives and to verify the complementarity of these objectives, we conduct a fine-grained ablation study and the results are presented in Table \ref{ablation}. It is observed that, in CALVIN D$\rightarrow$D, removing future RGB-D prediction and 3D flow prediction reduces the average number of solved tasks by 0.07 and 0.06, respectively, while removing current depth prediction only leads to a decrease by 0.04, indicating that the contribution of dynamics-related objectives is more profound. In addition, removing any of the three learning objectives increases the losses of the other two objectives in the validation set of CALVIN D$\rightarrow$D as well as the validation set of our real-world data. This confirms that the proposed self-supervised learning objectives can benefit from each other.

\begin{figure*}
    \centering
    \includegraphics[width=0.75\linewidth]{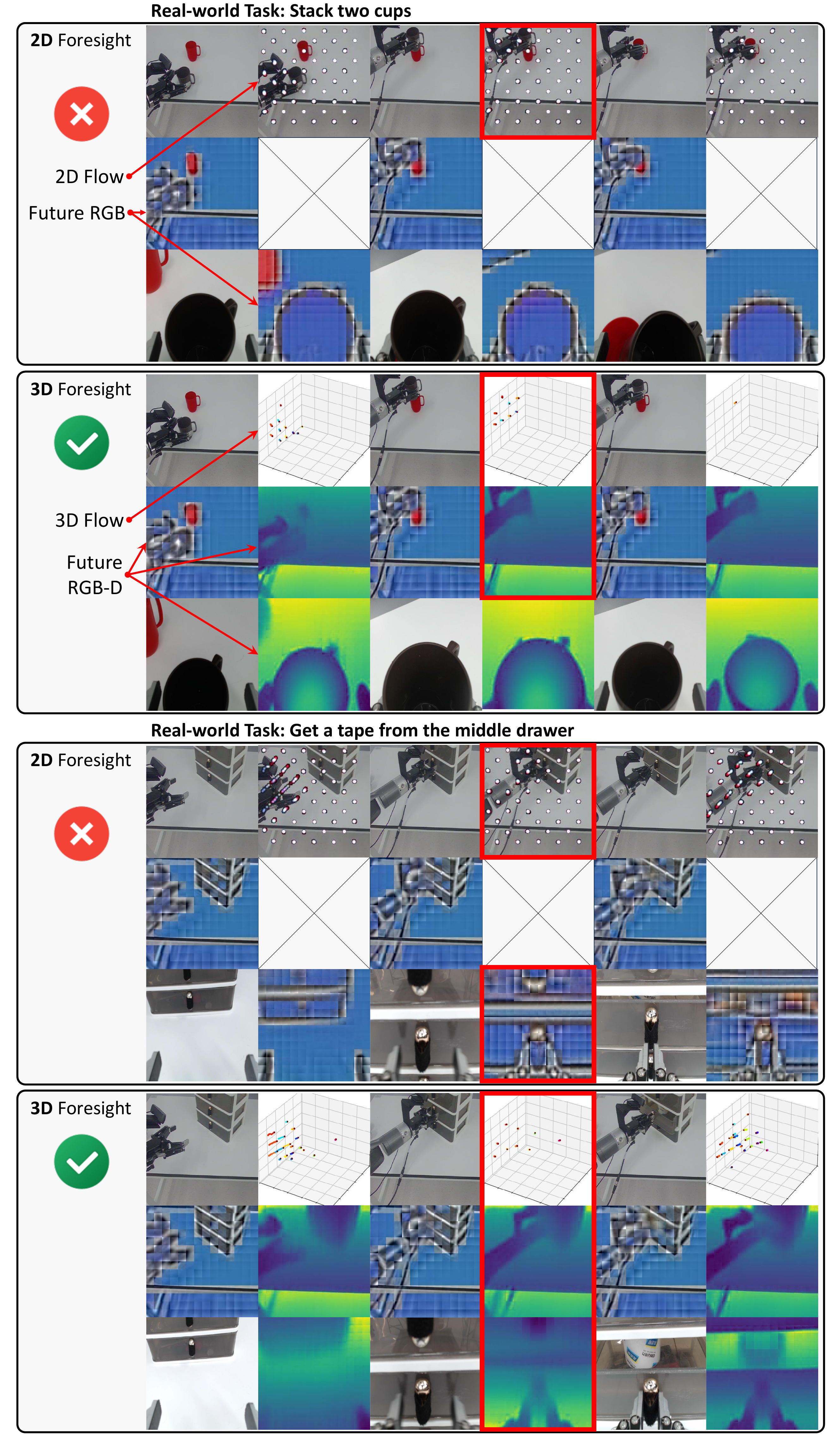}
    \caption{Visualization of two cases where the 3D Foresight policy succeeded but the 2D Foresight policy failed. Pixel-wise normalization is applied over future RGB during training, which is why the pictures here do not look like normal RGB.
}
   \label{fig:case}
\end{figure*}

\subsection{Real-world Evaluation and Case Study (Q4)}

We carry out real-world experiments to verify the effectiveness of our method in more complex and noisy scenarios. Two tasks (stack two cups, get a tape from the middle drawer) are considered, both of which require strong spatial awareness. For the first task (stack two cups), we design a special evaluation configuration, which involves two placement settings: 1) \textbf{Horizontal}, the cups are placed horizontally; 2) \textbf{Longitudinal}, the cups are placed longitudinally (see Fig. \ref{fig:env}). According to the evaluation results shown in Table \ref{realwolrd_result}, 3D foresight consistently improve policy performance throughout the tasks. And the performance gain is most prominent in the ``stack two cups'' task under the longitudinal setting, which involves the most noticeable depth-wise movement. This phenomenon aligns well with our hypothesis.

To qualitatively analyze the performance gain brought by 3D foresight, we take a deeper look into two cases where the 3D Foresight policy succeeded but the 2D Foresight policy failed. The rollouts along with the auxiliary predictions are visualized in Fig. \ref{fig:case}. In the first case, the 2D Foresight policy failed to accurately locate the position of the target cup (the wrist view was occluded and the model had to rely on distance perception over the main view) and released the held cup about 6 cm in front of the target cup. In contrast, the 3D Foresight policy successfully located the target cup, thanks to its stronger ability to perceive depth. In the second case, the policies had to determine the timing to grip the drawer handle based on the distance from the gripper to the handle. Different from the first case, the wrist view played a major role in the second case. And it turns out that the 3D Foresight policy does better at perceiving depth from the wrist view, as we explicitly teach that during training.

\section{CONCLUSION}

We explore a way towards 3D dynamics-aware robotic manipulation in this paper. To enhance manipulation policies with 3D foresight, a novel framework is introduced, which integrates 3D world modeling and policy learning through three auxiliary objectives — current depth estimation, future RGB-D prediction, and 3D flow prediction. Experiments across simulation and the real world demonstrate that policies can benefit much more from 3D foresight than from 2D foresight. Our ablation study reveals the complementarity of the proposed learning objectives and the major contribution of 3D dynamics-related objectives to performance. Our case study suggests that 3D foresight can provide manipulation policies with stronger spatial awareness which are critical for tasks that require distance perception or involve prominent depth-wise movement. A promising direction for future work is to explore more advanced 3D scene representations to further enhance the model's spatial reasoning capabilities.

\section{ACKNOWLEDGMENT}

This work is in part supported by the Guangzhou-HKUST(GZ) Joint Funding Program (2025A03J3656) , in part supported by the Guangdong Basic and Applied Basic Research Foundation (No. 2026A1515012291).

\bibliographystyle{IEEEtran}
\bibliography{references}


\end{document}